%% file: semeval2024.tex
\pdfoutput=1
\documentclass[11pt]{article}

\usepackage[final]{acl}
\usepackage{hyperref}
\usepackage{times}
\usepackage{latexsym}
\usepackage{multirow}
\usepackage{booktabs}
\usepackage{microtype}
\usepackage{multirow}
\usepackage{latexsym}
\usepackage{booktabs}
\usepackage{courier}
\usepackage{amsmath}
\usepackage{booktabs} % For better table formatting
\usepackage{siunitx} % For aligning numbers by decimal point
\usepackage{subfig}
\usepackage{placeins}
\usepackage{tikz}
\usepackage{hyperref}
\usepackage{cleveref}
\usepackage{adjustbox}
\usepackage{quoting}

\usepackage[T1]{fontenc}
\usepackage[utf8]{inputenc}
\usepackage{microtype}

\usepackage{inconsolata}

\title{UniBuc at SemEval-2024 Task 2: Tailored Prompting with Solar for Clinical NLI}

\author{
Marius Micluța-Câmpeanu$^{\clubsuit,\heartsuit,}$\thanks{These authors contributed equally to this work.},
\mbox{   }\mbox{   } Claudiu Creangă$^{\clubsuit,\heartsuit,}$\footnotemark[1]\\
\textbf{\mbox{   }\mbox{   }Ana-Maria Bucur$^{\clubsuit,\heartsuit}$,}
\textbf{\mbox{   }\mbox{   }Ana Sabina Uban$^{\spadesuit,\heartsuit}\textbf{,} \mbox{   }\mbox{   }$
Liviu P. Dinu$^{\spadesuit,\heartsuit} \mbox{   }\mbox{   }$}  
 \\ 
  $^\spadesuit$ Faculty of Mathematics and Computer Science
   \\
  $^\clubsuit$Interdisciplinary School of Doctoral Studies, $^\heartsuit$HLT Research Center \\
  University of Bucharest, Romania\\
  \small
{\tt marius.micluta-campeanu@unibuc.ro, claudiu.creanga@s.unibuc.ro}  \\
\small
  {\tt ana-maria.bucur@drd.unibuc.ro, \{auban, ldinu\}@fmi.unibuc.ro}\\  
 \vspace{-2mm}
}
\vspace{2mm}

\begin{document}
\maketitle
\begin{abstract}

This paper describes the approach of the UniBuc team in tackling the SemEval 2024 Task 2: Safe Biomedical Natural Language Inference for Clinical Trials. We used SOLAR Instruct, without any fine-tuning, while focusing on input manipulation and tailored prompting. By customizing prompts for individual CTR sections, in both zero-shot and few-shots settings, we managed to achieve a consistency score of 0.72, ranking 14th in the leaderboard. Our thorough error analysis revealed that our model has a tendency to take shortcuts and rely on simple heuristics, especially when dealing with semantic-preserving changes.

\end{abstract}

% - Tutorial Document: 

% \href{https://www.overleaf.com/read/hjsgxhwpzpws}{https://www.overleaf.com/read/hjsgxhwpzpws} [applied]

% - Template: 

% \href{https://www.overleaf.com/read/wmvfcjmfcqtf}{https://www.overleaf.com/read/wmvfcjmfcqtf}

% am impartit articolul in sectiuni. in functie de cat spatiu ocupa fiecare, le putem condensa la final

\section{Introduction}
\input{sections/introduction}

\section{Related Work}
\input{sections/related-work}
% Background

\section{Data and Task Description}
\input{sections/data}

\section{Methodology}
%if anyone has a better name for this, it can be changed
\input{sections/methodology}

\section{Results}
\input{sections/results}

\section{Conclusions and Future Work}
\input{sections/conclusion}

% \clearpage

\section*{Acknowledgements}
This work was partially supported by a grant on Machine Reading Comprehension from Accenture Labs and by the POCIDIF project in Action 1.2. ``Romanian Hub for Artificial Intelligence''.

\bibliography{semeval}

\appendix

% \section{Appendix}
% \label{sec:appendix}

\section{Additional examples}
\label{sec:examples}

\subsection{Summary mismatch example}\label{sec:mismatch-example}

The Results section of the CTR with ID ``NCT00467844'' is the following:
\begin{quoting}[rightmargin=0em]
    Outcome Measurement:   The Efficacy of GTx-024 on Total Body Lean Mass.  Change in total body lean mass as measured by dual energy x-ray absorptiometry (DEXA)from baseline to 4 months.  Time frame: Baseline to Four Months.
    
    Results 1:   Arm/Group Title: GTx-024 1 mg  Arm/Group Description: [Not Specified]  Overall Number of Participants Analyzed: 32  Median (Full Range)  Unit of Measure: kg  1.55        (-2.06 to 12.64)
    
    Results 2:   Arm/Group Title: GTx-024 3 mg  Arm/Group Description: [Not Specified]  Overall Number of Participants Analyzed: 34  Median (Full Range)  Unit of Measure: kg  0.98        (-4.84 to 11.54)
\end{quoting}

The summary is conditioned on the following hypothesis: ``all patients treated with gtx-024 1mg in the primary trial gained lean body mass over a 10 year period''. However, other hypotheses are concerned with other quantities: ``at least one patient treated with GTx-024 1mg in the primary trial gained over 10 kilos of Lean body Mass''. For the latter, our summary is misleading because the ``10 kilos'' information is missing. This could be mitigated by refraining to summarize short sections.

\section{Prompt templates}
\label{sec:prompts}
\input{sections/prompts}

\section{Infrastructure}
\label{sec:infrastructure}
\input{sections/infrastructure}

\end{document}

%% file: sections/introduction.tex
Clinical trials are prospective studies that aim to compare the effectiveness of an intervention against a control group in clinical patients \cite{friedman2015fundamentals}. ClinicalTrials.gov\footnote{\href{https://clinicaltrials.gov/}{https://clinicaltrials.gov/}} hosts more than 480,000 clinical trials, making it challenging to analyze and extract information from them manually. Natural language inference has emerged as a valuable tool for interpreting evidence from clinical trials \cite{jullien-etal-2023-nli4ct}.

% Clinical trials are prospective studies that aim to compare the effectiveness of an intervention (therapeutic drugs, vaccines, procedures, devices, etc.) against a control group in clinical patients \cite{friedman2015fundamentals}. These trials are important in determining which interventions are safe and effective, the incidence of adverse effects, and in developing personalized medicine practices. ClinicalTrials.gov\footnote{\href{https://clinicaltrials.gov/}{https://clinicaltrials.gov/}} hosts more than 480,000 clinical trials, making it challenging to analyze and extract information from them manually. Additionally, clinical contradictions are often found in medical corpora due to a large amount of newly developed studies \cite{makhervaks2023clinical}. To address these issues, various computational approaches have been proposed to automatically extract information \cite{kiritchenko2010exact} and to interpret evidence from clinical trials using natural language inference \cite{jullien-etal-2023-nli4ct}. 

The second task of SemEval 2024 focuses on improving the understanding of clinical trial data through the second edition of NLI4CT (Natural Language Inference for Clinical Trials) \cite{jullien-etal-2024-semeval}. This challenge is specifically designed to test the natural language inference capabilities of large language models (LLMs) and their ability to understand clinical text. The data used for the challenge was carefully annotated by clinical domain experts, and semantic interventions were performed to evaluate the safety and robustness of the models. 
%Submitted systems are evaluated using metrics designed to measure the robustness of their predictions, including consistency and faithfulness, in addition to the standard F1-score.

We employed LLMs, achieving the best results with SOLAR-Instruct, and focused on two key strategies:
\begin{itemize}
   \item  \textbf{Targeted Summarization}: Summarizing both CTRs and the hypothesis (retaining only the first 'trial' sentence) aided the model's focus on essential information.
   \item  \textbf{Tailored Prompting}: We used both zero-shot and two-shot prompting, tailoring prompts to individual CTR sections for optimal results.
\end{itemize}

We were surprised to find that causal models significantly outperform masked language models on this type of task. This is probably because the task requires reasoning capabilities that BERT-based models do not have. Our model's biggest challenges were with numerical reasoning and rephrasing (discussed in Section \ref{sec:discussion}), but despite those, we still secured the 14th place (out of 32) in the leaderboard. We make our code publicly available on GitHub\footnote{\url{https://github.com/ClaudiuCreanga/sem-eval-2024-task-2}}.

%% file: sections/related-work.tex
%Recent work on natural language inference for clinical trials studies a directly related problem to that of entailment which is addressed by this shared task: detection of clinical contradictions in publications from the medical domain \cite{makhervaks2023clinical}.  

% In the field of processing clinical trial data and assisting clinical trial development using language models, recent studies \cite{wang-etal-2023-autotrial,datta2024autocriteria} address the tasks of automatically assessing eligibility for clinical trials in an explainable manner using LLMs.

% In relation to the dominant methodologies used in recent approaches for this task (based on LLMs), including our own, and the suitability and safety of using LLMs for the medical domain, \citet{pal2023med} assess the prevalence of hallucinations in medical texts generated using various generative models. \citet{feng2023chard} assess the capabilities of LLMs to reason about medical data as a potential method to automatically generate explanations about health-related information. 
% The previous edition of the NLI4CT shared task at SemEval 2023 \cite{jullien-etal-2023-semeval} included a similar entailment task as well as an evidence selection task. Most submitted systems used pretrained language models, the majority of which were discriminative, with some generative LLM approaches. A small minority of approaches use ontologies and rule-based algorithms. 

Recent work on clinical trial analysis includes detecting contradictions in medical publications \cite{makhervaks2023clinical}, automating eligibility assessment with LLMs \cite{wang-etal-2023-autotrial,datta2024autocriteria}, and assessing model hallucinations and reasoning capabilities in healthcare settings \cite{pal2023med,feng2023chard}. The previous SemEval NLI4CT task \cite{jullien-etal-2023-semeval} included a similar entailment task, with most submissions leveraging pre-trained language models. A small minority of approaches used ontologies and rule-based algorithms. 

Few of the language model-based approaches include preprocessing of the data prior to using it as input to the models. In our approach for the current task, we also attempt solutions based on both discriminative and generative language models, and in addition perform preprocessing of the input clinical trial data before feeding it to the models, such as summarization. While most of the best performing systems in last year's task employ some in-domain pre-training on medical data, we find that, from the models we experimented with, general LLMs perform as well as or better than medical ones. This could be explained by their larger size and instruction tuning techniques, but also because of the recent advances in general LLMs. However, we do not perform an exhaustive comparison of the two kinds of models (we use few medical LLMs in our experiments), so a definitive conclusion can not be drawn on this comparison.

The NLI4CT dataset \cite{jullien-etal-2023-nli4ct} is a unique benchmark dataset for Natural Language Inference (NLI) in the clinical domain that contains data from Clinical Trial Reports (CTRs). In contrast, the MedNLI \cite{romanov2018lessons} dataset is another benchmark dataset for NLI in the clinical domain, but it only contains clinical notes. To ensure that NLI models are robust and safe, the organizers of the NLI4CT task perform semantic-preserving and semantic-altering interventions of the hypotheses. According to \citet{jullien-etal-2023-nli4ct}, NLI models for clinical trials require not only biomedical reasoning but also numerical reasoning capabilities, as CTRs contain a large amount of quantitative information. In this regard, we conduct experiments using SOLAR 10B \cite{kim2023solar}, which was trained on question-answer pairs from the mathematical domain to enhance its mathematical capabilities.

%% file: sections/data.tex
The data used for this task is comprised of 1,000 breast cancer CTRs collected from ClinicalTrials.gov with 24,000 entailment relations annotated by clinical domain experts \cite{jullien-etal-2023-nli4ct}. Each CTR is comprised of 4 sections: eligibility criteria, intervention, results and adverse events.

% \noindent \textbf{Eligibility criteria:} A set of requirements for the inclusion or exclusion of patients in a clinical trial.

% \noindent \textbf{Intervention:} Information on the type of treatment, its dosage, frequency, and duration.

% \noindent \textbf{Results:} Outcome measures, units, and the results of the clinical trial.

% \noindent \textbf{Adverse Events:} Signs and symptoms experienced by the participants during the clinical trial.

Each sample from the NLI4CT dataset consists of the CTR premise (one of the 4 sections of the CTR), a statement, and an entailment label (Entailment or Contradiction). The premise can refer to only one CTR in the \textit{Single} type instance, or to a primary and a secondary trial in the \textit{Comparison} type. 
%As opposed to last year's shared task \cite{jullien-etal-2023-semeval} in which participants had to predict the entailment relation and retrieve evidence for the prediction,
The purpose of the current task is to evaluate the consistency of models and their ability to perform faithful reasoning \cite{jullien-etal-2024-semeval}. For this purpose, different semantic altering 
(Contradiction Rephrasing and Numerical Contradiction) 
or semantic preserving interventions 
(Paraphrase, Numerical Paraphrase and Definition) 
have been conducted on the evaluation data. The NLI4CT dataset consists of 1,700 entailment relations in the training split, 200 in dev, and 5,500 in the test split.

% These interventions are detailed below:

% \noindent \textbf{Paraphrase} (e.g., ``Nobody taking part in the primary trial suffered a UTI" -> ``there were no occurrences of urinary tract infections among the participants of the primary clinical trial")

% \noindent \textbf{Contradiction Rephrasing} (e.g., ``There was at least 1 case of infection in both the primary trial and the secondary trial" -> ``there were no reported infections in the primary clinical trial nor the secondary clinical trial")

% \noindent \textbf{Numerical Paraphrase} (e.g., ``365 days"  -> ``52 weeks")

% \noindent \textbf{Numerical Contradiction} (e.g.,  ``3 years" -> ``8 years")

% \noindent \textbf{Definition} appended to the original statement (e.g., “gingival hemangioma is a hemangioma arising from the gums.”)

% (Figure \ref{tab:dataset}).

% \begin{table}[hbt!]
%     \centering
%     \begin{adjustbox}{width=0.35\textwidth}
%     \begin{tabular}{ccc}
%     \toprule
%      \textbf{Split}&  \textbf{Intervention}& \textbf{Samples}\\
%      \midrule
%      Train&  -& 1,700\\
%      \specialrule{.01em}{0em}{0em}
%      Dev&  -& 200\\
%      \specialrule{.01em}{0em}{0em}
%      \multirow{ 6}{*}{Test}&  -& 500\\
%      &  Paraphrase& 1,500\\
%      & Contradiction&1,500\\
%      & Definition&1,500\\
%      & Numerical Paraphrase&224\\
%      & Numerical Contradiction&276\\
%      \bottomrule
%     \end{tabular}
%     \end{adjustbox}
%     \caption{Number of samples in NLI4CT dataset.}
%     \label{tab:dataset}
% \end{table}

To assess the performance of the models in the shared task, two metrics have been proposed: \textbf{Faithfulness} and \textbf{Consistency} \cite{jullien-etal-2024-semeval}, besides F1-score. Faithfulness measures the model's ability to adjust its predictions accurately after a semantic-altering intervention. Consistency measures the capacity of the system to predict the same label for both the original and contrast statements, in the case of semantic-preserving interventions.

%% file: sections/methodology.tex
In this section, we present the different approaches used to predict the entailment relations.

\subsection{Pre-trained Masked Language Models}

Our first approach for the task of entailment relations prediction is using pre-trained models. Previous research has shown that domain-specific pre-training is beneficial for biomedical tasks \cite{Tinn2022,romanov2018lessons}.

We use pre-trained models, such as PubMedBERT \cite{gu2021domain}, XLM-RoBERTa \cite{conneau2020unsupervised}, Deberta V3 Large \cite{he2021deberta}, and fine-tune them on the training data. The fine-tuning process is done in 2 steps, as suggested in \cite{sun2020finetune}. Firstly, we stack a fully connected layer on top of the pre-trained model and train it for 4 epochs while the weights of the pre-trained model are frozen with a learning rate of $10^{-3}$. For the second step, we train the fully connected layer along with the last layer of the pre-trained model at a lower learning rate of $2*10^{-4}$ for one more epoch.

Inspired by the approach taken by \citet{pahwa2023bphigh}, who demonstrated that the performance of fine-tuned cross-encoders is comparable to that of GPT-3.5 in the few-shot setting, we experiment with the sentence-transformer model BioBERT\footnote{\href{https://huggingface.co/pritamdeka/BioBERT-mnli-snli-scinli-scitail-mednli-stsb}{pritamdeka/BioBERT-mnli-snli-scinli-scitail-mednli-stsb}} trained on 6 benchmark NLI datasets \cite{deka2022evidence} for sentence similarity tasks. We train a cross-encoder model using the sentence embeddings from BioBERT on the NLI4CT train data for sentence-pair classification for 20 epochs.

We also utilized SciFive \cite{phan2021scifive}, a T5 model designed for biomedical literature-related tasks. This model was pre-trained on a large corpus of PubMed abstracts and PubMed Central full-text articles from biomedical and life sciences domains. It achieved state-of-the-art results on the MedNLI benchmark dataset \cite{romanov2018lessons}. We used the SciFive model trained on MedNLI\footnote{\href{https://huggingface.co/razent/SciFive-large-Pubmed\_PMC-MedNLI}{razent/SciFive-large-Pubmed\_PMC-MedNLI}} in zero-shot setting to predict the entailment relations for the NLI4CT data.

\subsection{Large Language Models} \label{sec:llms}

LLMs have achieved promising results in biomedical tasks, such as named entity recognition, relation extraction, text classification, and question answering \cite{jahan2023comprehensive}. We conducted experiments using LLaMa-2 7B\footnote{\href{https://huggingface.co/meta-llama/Llama-2-7b}{meta-llama/Llama-2-7b}} \cite{touvron2023llama}, Mistral 7B\footnote{\href{https://huggingface.co/mistralai/Mistral-7B-Instruct-v0.2}{mistralai/Mistral-7B-Instruct-v0.2}} \cite{jiang2023mistral} and SOLAR \cite{kim2023solar}, which have shown promising results in various language tasks. LLaMa-2 is a competitive model that has performed well across multiple benchmarks such as commonsense reasoning, word knowledge, and reading comprehension. Mistral, on the other hand, has surpassed LLaMa-2 in all the tested benchmarks. We choose SOLAR-Instruct since it is a state-of-the-art model that is instruction-tuned specifically to have improved mathematical capabilities, with rephrased examples using a similar process to MetaMath \cite{yu2023metamath}. The team behind SOLAR developed a unique scaling technique, called depth up-scaling (DUS), which combines architectural changes with continued pretraining and obtained better results than larger models like Mixtral \cite{kim2023solar}. 

LLaMa-2 and Mistral were evaluated only in zero-shot settings on the NLI4CT test data. The same prompt was used in the experiments, regardless of the section the statement was referring to and it is presented in \Cref{sec:prompts}. While the experiments from LLaMa-2 and Mistral were using the entire sections of CTRs and statements, we had a different approach for SOLAR, which involved CTRs summarization. We expand on the methodology below.

\textbf{Section content summarization approach.} This approach consists of two stages. First, we summarize the text of each section to reduce the number of tokens.
Next, we perform the classification task on the shortened text using two-shot prompting with examples from the training set.
Both stages of this pipeline use the SOLAR 10.7B Instruct v1.0 model \citep{kim2023solar} in GGUF format\footnote{\href{https://huggingface.co/TheBloke/SOLAR-10.7B-Instruct-v1.0-GGUF}{TheBloke/SOLAR-10.7B-Instruct-v1.0-GGUF}} with 5-bit quantization (Q5\_K\_M) using \href{https://github.com/ggerganov/llama.cpp/tree/c9b316c78fba31e65879a2ec91cbafd341b88cce}{llama.cpp}.

% found using huggingface llm leaderboard: https://huggingface.co/spaces/HuggingFaceH4/open_llm_leaderboard

\textbf{CTR summarization.} We summarize the sections of each CTR in the evaluated datasets (train, development, test) to a maximum of 350 tokens.
The main reason for performing summarization is to provide the model with a shorter context and a task that is easier to solve. To validate this statement, we perform preliminary runs on the development set on Single CTRs for the Results section. We obtain an F1-score of 0.55 on single results CTR without a summary, while we are able to reach 0.63-0.72 F1-score with the summarization approach.

Another motivating factor is the time required to run the inference in order to perform multiple experiments. Full CTR inference for one example can take up to 20-30 seconds. Conversely, shortened CTRs are evaluated in roughly 5 seconds. As generating summaries is a one-time cost, the time difference is compensated for after a few iterations.
Evaluating on CTR summaries instead of the full CTR allowed us to dedicate more time in designing and refining prompts used for the entailment task. For these reasons, we do not conduct additional experiments on full CTRs.

The obvious drawback of summarization is the risk of discarding essential information. 
In the initial experiments for this approach, we tried to mitigate this by conditioning the summary to be related to the
hypothesis statement. Unfortunately, this strategy caused the model to include the statement in the summary and at times
even output contradictory phrases. Moreover, we did not try to continue with contextualized summaries because it would
require generating a new summary for each statement instead of a summary for each CTR section. Given that the evaluation
relies on using the same section with multiple statements, we need to generate only one summary per section if the summary
does not depend on the hypothesis.

\textbf{Inference.} We solve the entailment task through zero-shot and two-shot prompting.
As some statements might contain irrelevant sentences, we only keep the first sentence that contains the word ``trial''. We are aware that it is not an ideal approach and we lose important information in some examples where 2 or more sentences are crucial for the task. We found that the model works better if we keep only the ``trial'' sentence rather than not performing this step at all.
We also tried a more robust approach by asking the model to extract the relevant sentences from the hypothesis. While this tackles the issues encountered by our simple heuristic mentioned before, we limit our system setup to sentence splitting since it is significantly faster.

We limit the output grammar of the model to only ``Yes'' and ``No'' to ease processing. For each of the four sections and example types (Single or Comparison), we apply different
prompts for summarization and evaluation.
%summarization prompts and evaluation prompts.
The final prompt templates are listed in \Cref{sec:prompts}. % they take up a lot of space
The advantage of this approach is that we can analyze and tune the prompts independently for each section,
without running the inference step for the whole dataset.

% this is pretty much everything regarding system description
% the rest would only be failed/discarded attempts

% this would go in discussions or scrapped due to space??
%At first, we used a single prompt for evaluating all sections, reaching an F1-score of $\approx$ 74\%
% on the development set. 

%% file: sections/results.tex
% We would like to emphasize the importance of analysis of your system’s behavior beyond the overall scores. There will be a best paper award to recognize system papers with strong analysis.

\begin{table}[hbt!]
  \begin{adjustbox}{width=0.5\textwidth}
  \begin{tabular}{lcccc}
    \toprule
      \textbf{Model} & \textbf{Setting} & \textbf{F1} & \textbf{Faithfulness} & \textbf{Consistency} \\
    \midrule
    \texttt PubMedBERT & fine-tuned & 0.63 & 0.53 & 0.62 \\
    \texttt XLM-RoBERTa & fine-tuned & 0.63 & 0.55 & 0.63 \\
    \texttt Deberta V3 Large & fine-tuned & 0.63 & 0.54 & 0.62 \\
    \texttt BioBERT & fine-tuned & 0.65 & 0.52 & 0.63 \\    
    \midrule
    \texttt SciFive Pubmed PMC &zero-shot & 0.56 & 0.61 & 0.56 \\
    \texttt Llama-2 7B & zero-shot & 0.65 & 0.19 & 0.44 \\
    \texttt Mistral 7B & zero-shot & 0.65 & 0.18 & 0.44 \\
    \texttt Mistral 7B Instruct-0.2 & zero-shot & \textbf{0.72} &  0.68 & \underline{0.66} \\
    \texttt SOLAR 10B * & {zero-shot} & 0.63 & \textbf{0.90} & \textbf{0.72} \\
    \midrule
    \texttt SOLAR 10B & {few-shot}  & \underline{0.71} & \underline{0.83} & \textbf{0.72} \\
    \bottomrule
  \end{tabular}
  \end{adjustbox}
    \caption{Results of our submissions for the SemEval-2024 Task 2: Safe Biomedical Natural Language Inference for Clinical Trials. Best results are presented in \textbf{bold}, and second-best results are presented in \underline{underline}. Results with asterisk (*) were not submitted.}
      \label{tab:official-results}
\end{table}

The official results of our team can be found in Table \ref{tab:official-results}. Our results indicate that using pre-trained models on clinical text does not significantly improve the performance on this particular task, despite research confirming that domain-specific language model pre-training can be beneficial for other biomedical tasks \cite{Tinn2022}. In line with last year's findings \citep{jullien-etal-2023-semeval}, we observe that instruction-tuned models pre-trained on generic datasets perform better than discriminative models pre-trained on biomedical datasets. With respect to LLaMa-2 and Mistral models used in zero-shot settings, they achieve high F1 scores of 0.65 and 0.72. However, these models are not robust enough and achieve low performance on Faithfulness and Consistency metrics, which are the metrics the organizers focused on. We further expand on the results of our best-performing model, SOLAR.

% Control set: comparativ cu anul trecut, F1 ar fi în top 5, însă cu un model mult mai mic (fiind cuantizat)
% și folosind 2-shot, fără fine-tuning
% SOLAR este < 1/2 din Flan-T5; nu merge comparat cu Flan-T5 de anul trecut pt că avem rezultate pe test set
% doar cu few-shot, iar ei fac doar 0-shot

\textbf{Control set.} We obtain an F1-score of 0.71, with the highest score for comparisons of adverse events (0.79 F1)
and the lowest score for comparisons of interventions (0.62 F1). Our team reaches the {16th} place out of 32 participants in the official leaderboard. Compared to last year, we would be ranked on {5th} place while using little to no training data and modest computational resources. Similar to last year, we report a higher Recall (0.73) than Precision (0.70).

\textbf{Contrast set.} Our system achieves a Faithfulness score of 0.83 (10th place out of 32 teams) and a Consistency
score of 0.72 (14th place out of 32 teams). This shows that the model is more reliable when dealing with semantic-altering transformations compared to semantic-preserving changes, with only one CTR section having a faithfulness score below 0.79 (eligibility comparisons - 0.71). We observe a similar behavior in 22 other submissions where faithfulness
is higher than consistency.

\section{Error Analysis and Discussion}
\label{sec:discussion}
% \subsection{??}

In this section, we analyze the types of errors made by the SOLAR model according to the provided metrics based on each intervention target. % We omit from this analysis systematic errors occurring due to implementation defects.

\textbf{Definition interventions.} These interventions simply append a sentence to the statement. The model is capable of extracting the relevant sentence containing references to clinical trials if asked explicitly through a separate pre-processing step, but this incurs an additional runtime cost. Ultimately, we tackle this issue with a simple heuristic (see the inference details in \ref{sec:llms}).
% the heuristic was not implemented properly, which led to bogus splits: e.g. 0.3cm -> 3cm

% Definition interventions: aici modelul s-a descurcat bine
% TODO: de văzut impactul sumarizării și 0-shot vs few-shot; de căutat 1-2 exemple pe care nu s-a descurcat
% Definitions_Consistency:  0.7046666666666667
% Definitions_F1:  0.6955326460481099
% Definitions_Recall:  0.7177304964539007
% Definitions_Precision:  0.6746666666666666

% de văzut pe fiecare secțiune în parte
% cel mai prost a dat pe comparison_eligibility, comparison_intervention și single_adverse_events

\textbf{Numerical interventions.} Even though the SOLAR model has been tuned for mathematical reasoning, we identified several shortcomings. The model performs poorly with measurement units that express the same quantity in different ways. It appears to understand the meaning of symbols (e.g. ``positive'' instead of ``+''), but if domain-specific acronyms are lowercase instead of uppercase (e.g. ``hr'', ``her2'', ``mcs'', ``pdr''), the prediction changes.

Another interesting example is related to how entailment is affected when numbers have similar semantic meanings in other contexts. In one of the intervention trials, it is specified that a treatment cycle takes 21 days. The statement asks whether the trial has a 30-day cycle, with the model classifying it as entailment. Slightly tweaking the number in the statement reveals that this is not regarded as entailment if the values do not match semantically: only the pairs 21-28, 21-30 and 21-31 are considered entailment; the other pairs from 21-26 to 21-34 are classified as contradiction. Nonetheless, if we change the number in the trial (21) instead of the statement, it always predicts an entailment, even when the value is nonsensical. This appears to be a drawback of long contexts as the issue only manifests with 2-shot prompting.

For numerical paraphrasing, we have 0.67 Consistency, 0.54 F1-score, 0.63 Recall and 0.47 Precision. Conversely, for numerical contradictions, the model obtains 0.80 Faithfulness, 0.85 Consistency, 0.91 F1, 1.0 Recall and 0.83 Precision\footnote{We adapted the evaluation script to use \texttt{pos\textunderscore label} as 0 for contradictions and numerical contradictions.}.

% Numerical_Para_Consistency:  0.6741071428571429
% Numerical_Cont_Faithfulness:  0.8070175438596491
% Numerical_Cont_Consistency:  0.8580246913580247
% Numerical_Para_F1:  0.5408805031446541
% Numerical_Para_Recall:  0.6323529411764706
% Numerical_Para_Precision:  0.4725274725274725
% Numerical_Cont_F1:  0.911242603550296            % cu pos_label=0
% Numerical_Cont_Recall:  1.0                      % cu pos_label=0
% Numerical_Cont_Precision:  0.8369565217391305    % cu pos_label=0

% numerical contradictions - erorile nu par să fie (doar) din cauza alterațiilor numerice
% cf confusion matrix nimerește majoritatea exemplelor, dar script-ul nu e adaptat corect:
%   contradictions au doar gold labels negative și f1_score e by default average='binary' cu pos_label=1,
%   dar datele au doar pos_label=0

\textbf{Paraphrasing and contradiction interventions.} We notice a similar behavior to the numerical interventions: very high scores for contradictions, but significantly lower results for rephrases. Thus, paraphrases have 0.70 Consistency, 0.69 F1, 0.72 Recall, and 0.66 Precision, while contradictions have 0.83 Faithfulness, 0.78 Consistency, 0.89 F1, 1.0 Recall, and 0.80 Precision.

Our results suggest the model is overly sensitive to any wording change between the hypothesis and premise, mistakenly interpreting them as conflicting. This also explains the perfect recall scores.

% Para_Consistency:  0.7033333333333334
% Cont_Faithfulness:  0.8346666666666667
% Cont_Consistency:  0.7813333333333333
% Para_F1:  0.6903270702853167
% Para_Recall:  0.7219796215429404
% Para_Precision:  0.6613333333333333
% Cont_F1:  0.8938053097345133    % cu pos_label=0
% Cont_Recall:  1.0               % cu pos_label=0
% Cont_Precision:  0.808          % cu pos_label=0

% \subsection{Discussion}
In the remainder of this section, we present a few high-level remarks related to our design decisions.

% - 2-shot în general ajută și e o îmbunătățire față de zero-shot, indiferent de configurație; cu 6-shot se degradează performanța, cel puțin cu modelele și prompts încercate de noi; task-ul este dificil: deși nu am atins limita maximă de 4096 tokens a modelului, modelul nu pare să se descurce cu mai multe exemple în același prompt
% dă mai prost cu few-shot decât cu zero-shot pe exemple lungi - eligibility comparisons

% \textbf{Few-shot prompting consistently improves results compared to zero-shot prompting.}
% We used two examples from the training set, an entailment and a contradiction. This approach improved the results on all available metrics, regardless of model, model size or summarization configuration. Adding more examples to few-shot prompting might further increase the results at the expense of slower inference speed. However, with 6-shot prompting, the performance degrades even if we do not reach the context limit of 4096 tokens. We assume this is caused by the complexity of the task and the model ``forgetting'' what task it must solve.
\textbf{Few-shot prompting might improve results compared to zero-shot prompting.}
We used two examples from the training set, an entailment and a contradiction. This approach improved the results on average on the development set in terms of F1, regardless of model, model size or summarization configuration. Adding more examples to few-shot prompting might further increase the results at the expense of slower inference speed. However, with 6-shot prompting, the performance degrades even if we do not reach the context limit of 4096 tokens. We assume this is caused by the complexity of the task and the model ``forgetting'' what task it must solve. We do not further explore 6-shot prompting because of poor preliminary performance and increased runtime on a subset of the development data.

On the test set, it appears that few-shot prompting degrades the results in terms of Faithfulness and Contradiction, although it improves paraphrasing scores. We argue that a higher Faithfulness score in itself does not imply better results because the model could simply predict more Contradictions when the input is altered.

\textbf{Instruction order is significant.} We observed the best results when the instructions were placed at the
start of the prompt, followed by the CTR (or CTR summary) and then the hypothesis. This strategy constantly provides
improved predictions across most of the evaluated prompts. There is a drop of 4 F1 percentage
points on the development set when the hypothesis is placed before the CTR summary in the prompt template. Separately, the predictions are also affected if the instruction is placed at the end of the prompt, after the hypothesis. Repeating the instruction before and
after the CTR-hypothesis pair is as effective as simply placing the instruction before the CTR text.

\textbf{Larger models obtain better results, but smaller models are still useful for prototyping.} We use the
3-bit quantization version of the same SOLAR model (Q3\_K\_M) to experiment with more prompts, taking advantage
of faster inference times. This approach has been very useful in designing the summarization prompts because the
summarization step is the most expensive one in terms of computational resources. We also experimented with hybrid
systems, where the summaries are generated with a smaller model and the inference task is done by a larger model.
The performance of the hybrid strategy is comparable to running the entire pipeline with the larger model.

% - nu pare să fie atât de important/semnificativ ce prompt de sumarizare folosim, cât timp verificăm că în rezumatul rezultat sunt păstrate informațiile necesare pentru rezolvarea task-ului; modelul pare să se descurce și cu rezumate cu abrevieri

\textbf{There are no hard constraints for summaries prompts, as long as they do not depend on the hypothesis.}
There appears to be no substantial difference in the final results when changing the prompt used to generate CTR
summaries. We apply some of the following restrictions in each summarization prompt: use abbreviations, avoid verbs,
use short sentences, be brief, maximum N words. Rephrasing the prompt does not seem to have a relevant impact for
this task. As previously mentioned, it is paramount that the summaries preserve the meaning of the original text. The absence of relevant information in summaries is a major source of errors in our system.
Unfortunately, for most sections the model was unable to create contextualized summaries conditioned by
the hypothesis without mentioning the hypothesis in the summary.

For the Results section with a single CTR, conditioning the summary on the hypothesis looked promising.
However, the next example from the test set confirms our concerns about this strategy, where a single summary for multiple hypotheses is not appropriate due to conditioning on the first hypothesis in the dataset. Our generated summary is:  ``{There is no
information in the given CTR report that relates to the statement about all patients treated with GTx-024 1mg
gaining lean body mass over a 10 year period}''. The model is distracted by the ``10 year period'' from one of the hypotheses, altering the original meaning completely, even though the word ``year'' does not appear anywhere in the initial CTR section. See \cref{sec:mismatch-example} for the full CTR text and associated hypotheses.

%% file: sections/conclusion.tex
% The best performing systems were ....

% cel mai important mi se pare faptul că modelul pare să prezică mai mult contradicții deoarece informația nu e repetată cuvânt cu cuvânt => trebuie dezvoltată în continuare evaluarea pe reformulări pt că acum modelul pare să fenteze evaluarea

In this paper, we presented our approach for the SemEval 2024 Task 2 \citep{jullien-etal-2024-semeval} aimed at understanding large language models behavior in clinical contexts. We explored several types of models and prompting techniques in order to determine whether fine-tuning is more feasible than zero-shot or few-shot prompting in a limited resource setting.

Our findings suggest that, while LLMs exhibit remarkable clinical NLI capabilities
% even without explicit medical pre-training or fine-tuning
at a surface level, the proposed metrics and interventions uncover a tendency of the models to take shortcuts
and rely on simple heuristics, especially when faced with semantic-preserving changes.
We intend to investigate further methods of evaluating the reliability of large language models in future work. 

To address the inherent weak numerical reasoning of our model (and all generative models), a promising strategy is to offload complex mathematical hypotheses to a specialized model like xVal \cite{Golkar}. This approach involves representing numbers as individual digits (e.g., 123 becomes ["1", "2", "3"]), replacing them with a generic [NUM] token, and scaling the token according to the original numerical value. Their results showed a 70-fold improvement over standard models. We can extend this strategy to create an ensemble where other weaknesses of our model (like rephrases) are offloaded to specialized models.

% - efortul computațional este mult mai mare pentru a îmbunătăți un prompt generic comparativ cu utilizarea de prompts specializate; considerăm că este o abordare necesară deoarece conținutul diferă destul de mult între secțiuni

% Ideas for improvements could include ...

% pt ce ziceam cu ce exemple ar trebui folosite, s-a făcut și se cheamă "demonstration selection" - find the best examples for few-shot prompting (a.k.a. in-context learning)
% https://arxiv.org/pdf/2301.00234.pdf

%% file: sections/prompts.tex
The prompt templates used to obtain the final leaderboard results for SOLAR are shown in tables \ref{tab:eval_prompt} and
\ref{tab:summary_prompt}.

We started with a single prompt template for all sections and summaries. When the results did not further improve,
we analyzed the F1-score of each section, shown in Tables \ref{tab:init_results_train} and \ref{tab:init_results_dev}.
Due to time constraints, we only create
summaries for the first 50 examples in the train set.

% aici nu sunt sigur, a fost oarecum intenționat din lipsă de timp
While the single CTR summaries for the results section depend on the hypothesis, due to an implementation choice, % defect (?),
all examples use the same CTR summary, regardless of the hypothesis (only the first hypothesis is used). We believe
that this issue is not essential, since all the hypotheses for a CTR focus on the same information.

Initial summary prompt: ``Instruction: You are given a clinical trial report. You must summarize the report. Use abbreviations. Be brief. Report: \{premise\}. Summary (max 350 words):''.

Initial evaluation prompt: ``Instruction: You are given a Clinical Trial Report and a hypothesis.
\#\#Report: \{premise\}. \#\#Hypothesis: \{hypothesis\}. \#\#Can the hypothesis be inferred from the report? Respond only with Yes or No. \#\#Response (Yes or No):''. This prompt was also used for our experiments with LLaMa-2 and Mistral.

% TODO: de actualizat cu noile rezultate, rerulez script-ul cu un prompt inițial și apoi cu prompt-urile din paper - complet pt dev și train (first 50) să fiu sigur că cifrele de mai jos se reproduc (ish) și să justifice de ce am umblat mai mult doar la unele prompts;
% pt test sper să fie suficient doar rerulat evaluarea, fără rezumate

\begin{table}[h]
    \centering
    \begin{tabular}{lr}
        Section                     & F1     \\
        \toprule
        Eligibility (Single)        & 0.6637 \\ % mail: 0.5050
        Eligibility (Comparison)    & 0.5274 \\ % mail: 0.5508
        Intervention (Single)       & 0.7306 \\ % mail: 0.7342
        Intervention (Comparison)   & 0.7751 \\ % mail: 0.7887
        Results (Single)            & 0.7141 \\ % mail: 0.7550
        Results (Comparison)        & 0.7525 \\ % mail: 0.6991
        Adverse events (Single)     & 0.7141 \\ % mail: 0.7141
        Adverse events (Comparison) & 0.6805 \\ % mail: 0.7177
    \end{tabular}
    \caption{Initial results for the train set (first 50 examples)}
    \label{tab:init_results_train}
\end{table}
% total: 0.7040

\begin{table}[ht]
    \centering
    \begin{tabular}{lr}
        Section                     & F1     \\
        \toprule
        Eligibility (Single)        & 0.8178 \\ % mail: 0.6757; nu e chiar ok, a dat prea bine... 0.7953 cu fewshot
        Eligibility (Comparison)    & 0.8285 \\ % mail: 0.5804; 0.9160 cu fewshot
        Intervention (Single)       & 0.7678 \\ % mail: 0.8461; 0.8461 cu fewshot pe init prompt
        Intervention (Comparison)   & 0.6000 \\ % mail: 0.6969; 0.8 cu fewshot
        Results (Single)            & 0.7368 \\ % mail: 0.8421; 0.7103 cu fewshot
        Results (Comparison)        & 0.7749 \\ % mail: 0.7777; 0.8328 cu fewshot
        Adverse events (Single)     & 0.5835 \\ % mail: 0.6862; 0.6190 cu fewshot
        Adverse events (Comparison) & 0.6969 \\ % mail: 0.6969; 0.8496 cu fewshot
    \end{tabular}
    \caption{Initial results for the development set}
    \label{tab:init_results_dev}
\end{table}
% inițial rezumatele erau ~128 tokens și făcute cu modelul q3, probabil (și) din cauza asta discrepanța la scor
% eventual nu mai includem tabelele init_results_train și init_results_dev
%
% total fără fewshot: 0.7346, ceea ce reproduce rezultatele inițiale unde aveam F1 ~73-74
% problema a fost că am plecat cu prompts de la 0 când am făcut prompts pe secțiuni
% total cu fewshot: 0.7748, deci saltul cel mai mare pare că a fost din faptul că am făcut fewshot, nu din prompts

\begin{table*}[htbp]
    \centering
    \begin{tabular}{lp{10.59cm}}
        Section (CTR type)              & Prompt \\
        \toprule
        Eligibility (Single)        & {Instruction: You are given clinical trial criteria and a statement that may or may not be contradictory. Regarding the inclusion and exclusion criteria, is the statement correct? Respond only with Yes or No.\newline\#\# Criteria: \{premise\}.\newline\#\# Statement: \{hypothesis\}.\newline\#\# Response (Yes or No):} \\
        \midrule
        Eligibility (Comparison)    & {Instruction: You are given clinical trial criteria for a primary and a secondary trial, and a statement. Regarding the inclusion and exclusion criteria, is the statement correct for each trial? Respond only with Yes or No.\newline\#\# Criteria: \{premise\}.\newline\#\# Statement: \{hypothesis\}.\newline\#\# Response (Yes or No):} \\
        \midrule
        Intervention (Single)       & {Instruction: You are given a CTR and a statement. Can the statement be deduced from the CTR? Focus on the interventions. Respond only with Yes or No.\newline\#\#CTR: \{premise\}.\newline\#\#Statement: \{hypothesis\}.\newline\#\#Response (Yes or No):} \\
        \midrule
        Intervention (Comparison)   & (same prompt template as single CTR for interventions) \\
        \midrule
        Results (Single)            & {Instruction: You are given the results of a CTR and a statement. Can the statement be deduced from the CTR in terms of number of participants, measures and results? Respond only with Yes or No.\newline\#\#CTR: \{premise\}.\newline\#\#Statement: \{hypothesis\}.\newline\#\#Response (Yes or No):} \\
        \midrule
        Results (Comparison)        & {Instruction: You are given the results of a CTR and a statement. Can the statement be deduced from the CTR? Respond only with Yes or No.\newline\#\#CTR: \{premise\}.\newline\#\#Statement: \{hypothesis\}.\newline\#\#Response (Yes or No):} \\
        \midrule
        Adverse events (Single)     & {Instruction: You are given a CTR and a statement. Regarding the adverse events, signs and symptoms observed in the CTR, is the statement correct? Respond only with Yes or No.\newline\#\#CTR: \{premise\}.\newline\#\#Statement: \{hypothesis\}.\newline\#\#Response (Yes or No):} \\
        \midrule
        Adverse events (Comparison) & {(same prompt template as single CTR for adverse events)}
    \end{tabular}
    \caption{Evaluation templates for each CTR section}
    \label{tab:eval_prompt}
\end{table*}

\begin{table*}[htbp]
    \centering
    \begin{tabular}{lp{10.59cm}}
        Section (CTR type)              & Prompt \\
        \toprule
        Eligibility (Single)        & {Instruction: You are given the eligibility criteria for a clinical trial report. You must summarize the report focusing on inclusion and exclusion criteria.\newline Report: \{premise\}.\newline Use short sentences. Summary: } \\
        \midrule
        Eligibility (Comparison)    & {(same prompt template as single CTR for eligibility)} \\
        \midrule
        Intervention (Single)       & {Instruction: You are given the intervention information for a clinical trial report. Each report contains 1-2 cohorts, which receive different treatments, or have different characteristics. You must summarize the report focusing on the type, dosage, frequency, and duration of treatments being studied.\newline Report: \{premise\}.\newline Use short sentences to group by cohort. Summary:} \\
        \midrule
        Intervention (Comparison)   & {Instruction: You are given the intervention information for a clinical trial report. Each report contains 1-2 cohorts, which receive different treatments, or have different characteristics. You must summarize the report focusing on the type, dosage, frequency, and duration of treatments being studied.\newline Report: \{premise\}.\newline Use short sentences to group by cohort and group by primary trial and secondary trial. Summary:} \\
        \midrule
        Results (Single)            & {Instruction: You are given the results of a CTR and a statement. Extract all the relevant information from the CTR that is related to the statement.\newline Report: \{premise\}. Statement: \{hypothesis\}.\newline Answer:} \\
        \midrule
        Results (Comparison)        & {Instruction: You are given the results of two clinical trials. Each trial contains 1-2 cohorts, which receive different treatments, or have different characteristics. You must summarize the report for each trial, focusing on number of participants, outcome measures, units, results.\newline Report: \{premise\}.\newline Use short sentences and keep all numeric values. Summary:} \\
        \midrule
        Adverse events (Single)     & {Instruction: You are given the adverse events of a clinical trial report. Each report contains 1-2 cohorts, which receive different treatments, or have different characteristics. You must summarize the report focusing on the adverse events, signs and symptoms observed in patients.\newline Report: \{premise\}.\newline Use short sentences. Summary:} \\
        \midrule
        Adverse events (Comparison) & {Instruction: You are given the adverse events of a clinical trial report. Each report contains 1-2 cohorts, which receive different treatments, or have different characteristics. You must summarize the report focusing on the adverse events, signs and symptoms observed in patients.\newline Report: \{premise\}.\newline Use short sentences and group by primary trial and secondary trial. Summary:}
    \end{tabular}
    \caption{Summarization templates for each CTR section. The results section (single CTR) is the only one for which summaries depend on the hypothesis due to lack of time to rerun the summaries for the test set.}
    \label{tab:summary_prompt}
\end{table*}

%% file: sections/infrastructure.tex
% We used the following infrastructure(s) for running our models:

% -  ... execution time ...

% For the section content summarization approach,
In terms of infrastructure, we use a system with 16 GB RAM and an NVIDIA GTX 1060 MQ GPU with 6 GB VRAM. Out of the 5500 examples on the test set, this method only requires generating summaries for 251 examples comprising different CTR sections. On the development set, we need to create 100 summaries for the 200 samples, making the summarization step more expensive in this regard.

The inference time for a summary varies with the length of the CTR section, with a minimum time of 30 seconds per sample and a total time of about 7 hours, but it should be noted that this stage is a one-time cost.
The inference time for one example is approximately 5 seconds, meaning that the evaluation on the test set takes roughly 9 hours, with the total time reaching 12 hours when applying 2-shot prompting, since the sequence length increases.